\documentclass[letterpaper]{article} 
\usepackage[preprint]{aaai2027} 
\usepackage[hyphens]{url} 
\usepackage{graphicx} 
\usepackage{natbib} 
\usepackage{caption} 
\usepackage{amsmath,amssymb}
\usepackage{booktabs}
\usepackage{placeins}

\title{ActFovea: Runtime Safeguarding for VLA Policies via Spatiotemporal Visual-Action Consistency}

\author{
    Wenda Yu\textsuperscript{\rm 1},
    Tianshi Wang\textsuperscript{\rm 1},
    Fengling Li\textsuperscript{\rm 2},
    Xin Li\textsuperscript{\rm 3},
    Jingjing Li\textsuperscript{\rm 4},
    Lei Zhu\textsuperscript{\rm 1}
}
\affiliations{
    \textsuperscript{\rm 1}Tongji University\\
    \textsuperscript{\rm 2}Mohamed bin Zayed University of Artificial Intelligence\\
    \textsuperscript{\rm 3}Shanghai Artificial Intelligence Laboratory\\
    \textsuperscript{\rm 4}University of Electronic Science and Technology of China\\
    yu\_wenda@126.com, tswang0116@163.com, fenglingli2023@gmail.com\\
    sankin0528@gmail.com, lijin117@yeah.net, leizhu0608@gmail.com
}

\begin{document}

\maketitle

\begin{abstract}
Vision-language-action (VLA) policies achieve strong performance in robotic manipulation but remain vulnerable to runtime disturbances that break the temporal alignment among visual observations, robot states, and executed actions. We introduce ActFovea, a plug-and-play safeguarding framework that detects and mitigates such failures without retraining or modifying the underlying VLA policy. ActFovea uses robot kinematics, proprioceptive states, and recent actions to construct action-conditioned foveated regions that retain contact-relevant areas and predicted motion corridors while suppressing task-irrelevant visual content. It detects runtime risks by evaluating whether visual motion and observation freshness remain consistent with geometric, proprioceptive, and action transitions. For recoverable disturbances, ActFovea constructs disturbance-specific candidate observations and accepts a recovery only after verifying the resulting action chunk. When stale or replayed observations make reliable recovery impossible, it invokes a bounded safe-failure procedure. In closed-loop evaluations of $\pi_0$ across multiple LIBERO suites, ActFovea increases success under localized visual overlays from 49.3\% to 90.3\%, closing 93.7\% of the gap to clean performance. It further improves success under action drift and visual delay by 7.0 and 9.8 percentage points, respectively, while preserving clean-task performance. Under frozen-observation replay, ActFovea triggers timely safe failure in all trials, with no unprotected failures. These results demonstrate that spatiotemporal visual-action consistency provides an effective basis for runtime safeguarding of VLA policies.
\end{abstract}

\section{Introduction}
\begin{figure}[!t]
    \centering
    \includegraphics[width=0.98\columnwidth]{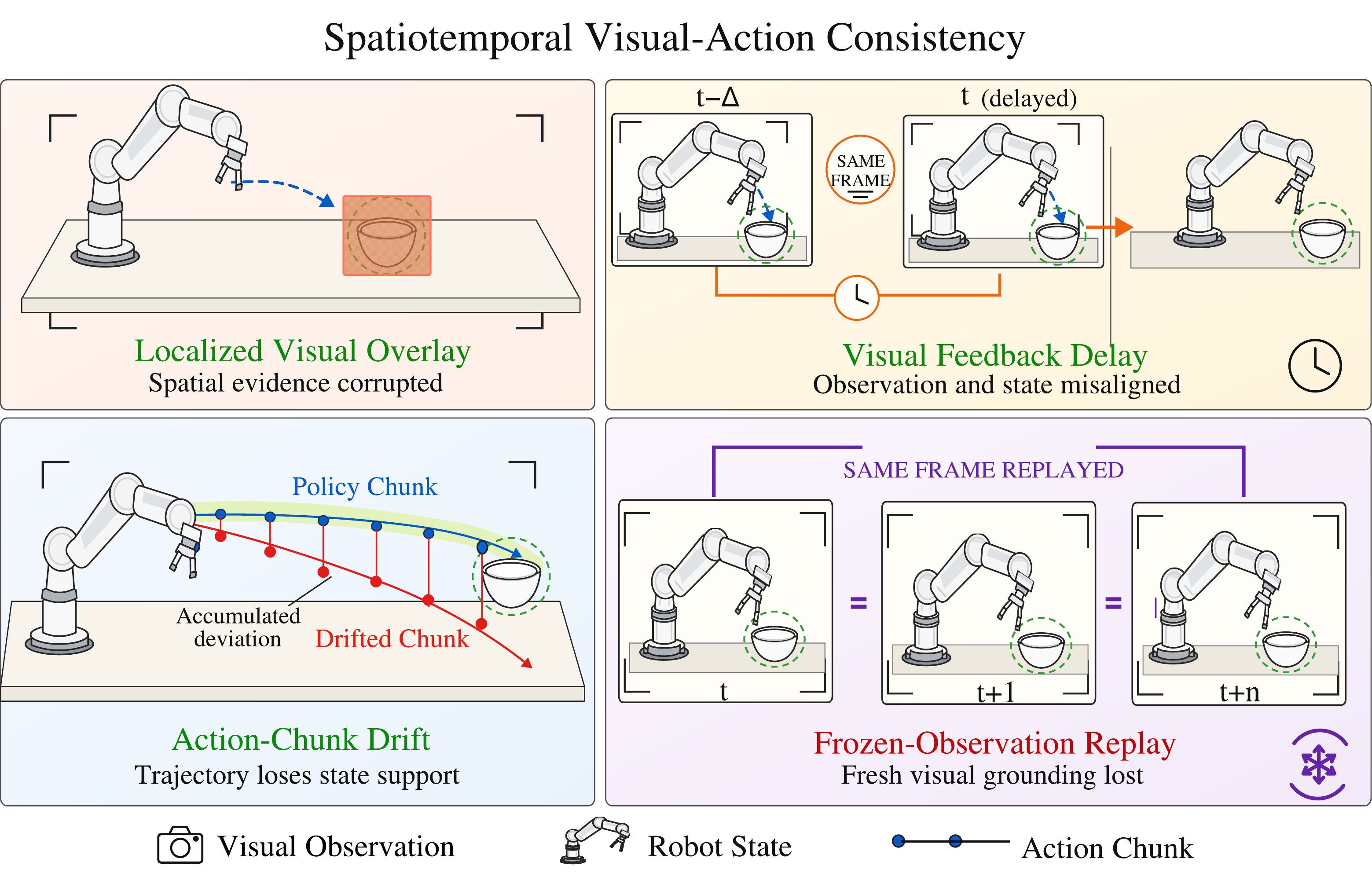}
    \caption{Four runtime disturbances that violate spatiotemporal visual-action consistency. Localized overlays corrupt spatial evidence, visual-feedback delay misaligns observations with the current robot state, action-chunk drift causes trajectories to lose state support, and frozen-observation replay eliminates fresh visual grounding.}
    \label{fig:introduction}
\end{figure}

Vision-language-action, or VLA, policies map visual observations and language instructions directly to robot actions, offering a scalable approach to general-purpose manipulation \cite{brohan2023rt2,kim2024openvla,black2024pi0}. Many current VLA policies predict action chunks, allowing multiple controls to be executed before the policy conditions on a new observation \cite{zhao2023act}. This design amortizes inference and promotes temporally coherent motion, but it also makes reliable control depend on continued alignment among visual observations, robot proprioception, and executed actions. When this alignment is broken, the policy may act on visual evidence that no longer corresponds to the robot’s current physical state. Figure~\ref{fig:introduction} illustrates how the disturbances considered in this work disrupt different components of this closed-loop relationship.

Recent studies have revealed vulnerabilities across these components. Visual attacks can modify task-relevant image evidence \cite{lu2025upa,lu2025phantom}, while SilentDrift exposes smooth intra-chunk action deviations \cite{xu2026silentdrift} and FreezeVLA shows that adversarial observations can induce persistent inaction \cite{wang2025freezevla}. Existing runtime safeguards provide complementary but largely specialized forms of protection. Control-barrier layers enforce explicit geometric constraints, while observation interventions reduce sensitivity to visual distractors \cite{hu2025vlsa,hancock2024byovla}.

These advances leave a deployment gap. A VLA policy may encounter spatial corruption, temporal misalignment, and action-trajectory drift during the same task, whereas existing safeguards typically target a particular disturbance family or apply a predetermined runtime constraint. A uniformly conservative response can unnecessarily suppress valid behavior, while treating every anomaly as recoverable can prolong control after reliable perceptual grounding has been lost. Runtime safeguarding must determine not only whether the closed loop has become inconsistent, but also whether a justified recovery remains possible.

Making this distinction is difficult because the expected visual transition depends on the current action, robot state, and task phase. A nearly static image may reflect either a legitimate pause or stale feedback, while a smooth action chunk may still deviate from the motion supported by the observed scene. These ambiguities must be resolved without access to reward signals, simulator object states, segmentation annotations, or prior knowledge of the disturbance type. The observable structure is spatiotemporal visual-action consistency: visual observations, proprioceptive transitions, and candidate actions should jointly describe a coherent physical evolution of the robot and its environment.

Based on this principle, we introduce ActFovea, a plug-and-play runtime safeguarding framework for VLA policies. ActFovea combines robot kinematics, proprioceptive transitions, visual history, and previous actions to construct action-conditioned foveated regions that preserve expected contact areas and predicted motion corridors. Unlike image-centered foveation, its visual support follows the anticipated interaction rather than the geometric center of the image. A consistency monitor identifies the active runtime risk and determines whether the available evidence is sufficient for recovery. For recoverable disturbances, ActFovea constructs a disturbance-conditioned bank of candidate observations and verifies the action chunk induced by each candidate before execution. When observations remain stale or replayed, ActFovea disables recovery and invokes a bounded motion-suppression procedure for safe failure. The underlying VLA policy remains frozen throughout. 

Across 40 LIBERO tasks with $\pi_0$, ActFovea recovers 93.7\% of the success-rate loss induced by localized visual overlays, improves performance under visual-feedback delay and smooth action-chunk drift, and preserves performance under undisturbed conditions. Under frozen-observation replay, it triggers timely safe failure in all evaluated trials, with no unprotected failures. Runtime comparisons and component ablations further quantify the contributions of disturbance-conditioned observation recovery and action-chunk verification across different forms of inconsistency.

In summary, our main contributions are as follows:
\begin{itemize}
    \item We formulate spatial corruption, temporal misalignment, action-trajectory drift, and observation replay as violations of spatiotemporal visual-action consistency jointly grounded in visual evidence, proprioceptive transitions, and action history.
    \item We develop a plug-and-play runtime safeguard that combines action-conditioned foveation, disturbance-conditioned observation recovery, action-chunk verification, and recoverability-aware safe failure without retraining or modifying the underlying VLA policy.
    \item We conduct controlled closed-loop evaluations across 40 LIBERO tasks, showing recovery from localized visual overlays, visual delay, and action drift, preserved clean performance, and timely safe failure under frozen-observation replay, supported by runtime comparisons and component ablations.
\end{itemize}

\begin{figure*}[!t]
    \centering
    \includegraphics[width=\textwidth]{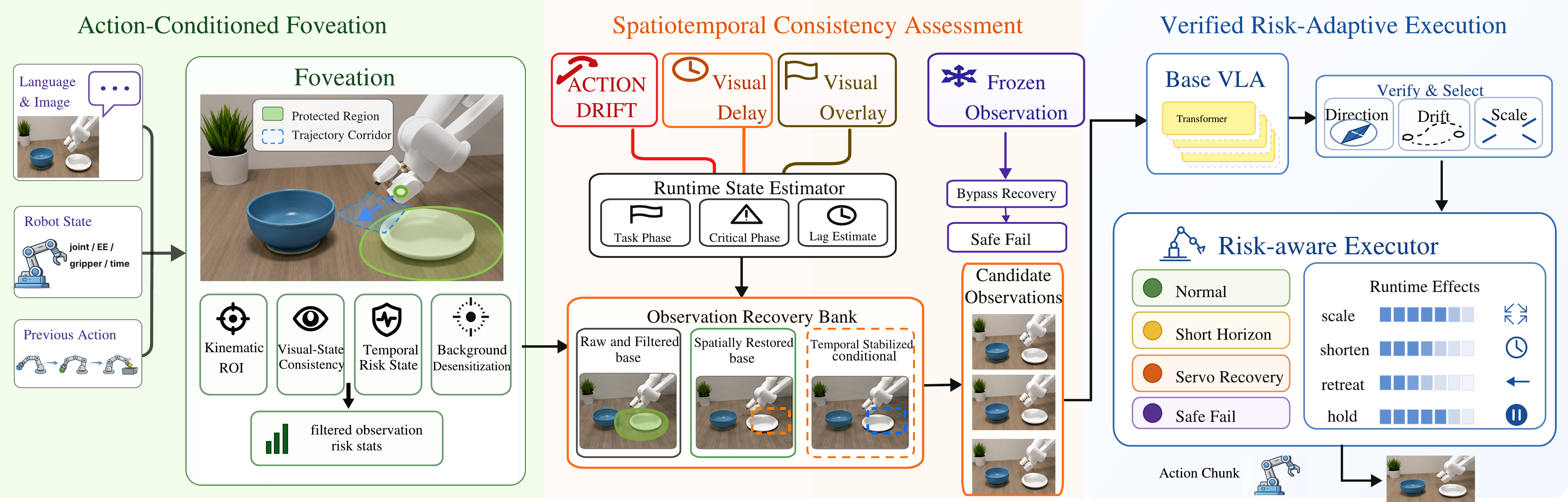}
    \caption{Overview of ActFovea. Action-conditioned foveation constructs
    interaction-centered preserve regions and consistency evidence from
    runtime observations, robot state, and action history. Recoverable spatial,
    temporal, and action-side inconsistencies enter a shared candidate,
    verification, and bounded-execution path. Confirmed frozen-observation
    replay instead bypasses recovery and triggers safe failure.}
    \label{fig:actfovea-framework}
\end{figure*}

\section{Related Work}

\subsection{VLA Robustness under Runtime Disturbances}
Vision-language-action policies couple visual and linguistic representations to robot control, enabling instruction-conditioned behavior across objects, tasks, and embodiments. Early systems such as RT-2 cast actions as tokens, while OpenVLA and Octo provide open generalist policies trained on diverse robot data, and $\pi_0$ generates continuous actions through flow matching \cite{brohan2023rt2,kim2024openvla,octomodelteam2024octo,black2024pi0}. Many contemporary policies predict action chunks rather than isolated controls, following the broader action-chunking paradigm introduced for visuomotor imitation learning \cite{zhao2023act}. Chunking amortizes policy inference and supports coherent motion, but it also places multiple executed controls between successive perceptual updates. Reliable deployment therefore depends not only on single-frame perception or individual actions, but also on their temporal alignment with the evolving robot state.

Recent security studies expose failures along each part of this closed loop. Attacks span prompt-induced control hijacking, sparse or transferable visual perturbations, and physical variations or sensor interference, while AttackVLA systematizes adversarial and backdoor evaluation across the VLA lifecycle \cite{jones2025adversarial,zhang2025advla,lu2025upa,liu2025evavla,lu2025phantom,li2025attackvla}. Backdoor studies further demonstrate persistent trigger-conditioned control deviations that preserve clean behavior \cite{zhou2025badvla}. SilentDrift identifies the intra-chunk accumulation of smooth action deviations, and FreezeVLA shows that adversarial observations can induce persistent inaction \cite{xu2026silentdrift,wang2025freezevla}. Although their threat models and adversary capabilities differ, these results reveal a common deployment failure: visual content, observation timing, proprioceptive evolution, and action trajectories can cease to describe the same physical transition. This literature primarily constructs or benchmarks individual attacks. ActFovea addresses the complementary defense problem, using representative visual, temporal, and action-space disturbances to study whether their shared spatiotemporal inconsistency can be detected and managed online.

\subsection{Runtime Safeguarding for Embodied Policies}
One line of work improves safety or robustness during training. SafeVLA formulates VLA alignment through constrained learning, whereas Phantom Menace uses adversarial training to improve robustness to physical sensor attacks \cite{zhang2025safevla,lu2025phantom}. Such approaches can internalize safety preferences or perturbation robustness when training data, optimization budgets, and parameter access are available. ActFovea considers a complementary deployment setting: the pretrained policy remains frozen, and safeguarding is performed through its observation-action interface.

Inference-time methods avoid retraining but enforce different runtime contracts. VLSA places a control-barrier-function layer after a VLA to enforce explicit collision constraints \cite{hu2025vlsa}. BYOVLA instead probes a policy's action sensitivity to image regions and minimally edits task-irrelevant visual distractors \cite{hancock2024byovla}. These approaches demonstrate the value of plug-and-play control and observation interventions, but respectively focus on geometric hazards and visual distractors. ActFovea targets loss of consistency across visual observations, robot state, and action chunks. It combines action-conditioned foveation with threat-conditioned observation candidates, verifies the resulting action chunks, and invokes bounded motion suppression when fresh evidence cannot be recovered. Thus, the contribution is a unified runtime safeguarding loop rather than a separate detector or correction rule for each disturbance.

\section{Method}

\subsection{Overview and Threat Model}
\label{sec:method-overview}

We consider a frozen VLA policy $\pi$ deployed in a closed-loop manipulation system. At query $t$, the observation $o_t$ contains RGB images $\{I_t^v\}_{v\in\mathcal V}$, available end-effector and joint/proprioceptive states, and an observation timestamp. Conditioned on $o_t$ and language instruction $l$, the policy predicts
\begin{equation}
A_t=\pi(o_t,l)=[a_{t,0},\ldots,a_{t,H-1}],
\label{eq:vla-policy}
\end{equation}
where $H$ is the chunk horizon. A prefix of length $h_t\leq H$ is committed before replanning. ActFovea interacts with $\pi$ through its observation-action interface and keeps the policy parameters frozen. It observes each action chunk before execution and maintains short histories of visual observations, proprioceptive states, and executed actions. Disturbances introduced after this interface are outside our threat model.

We study disturbances along three axes of the closed-loop transition. Spatial visual inconsistency is instantiated by a persistent localized visual overlay that alpha-blends a fixed pattern into the camera streams after an initial clean interval. Temporal visual inconsistency includes both multi-view visual feedback delay, which pairs an earlier image with current proprioception, and frozen-observation replay, which repeatedly supplies the same stale frame while the robot state evolves. Action-trajectory inconsistency is instantiated by smooth drift applied to the motion dimensions of a policy-generated chunk before ActFovea verifies it. These disturbances do not share an attack generator; they share the failure signature that visual content, proprioceptive transitions, and action evolution no longer support one coherent physical transition.

Figure~\ref{fig:actfovea-framework} summarizes the pipeline. ActFovea uses robot kinematics and action history as a reference for expected visual change and action evolution. It first constructs foveated observations and measures visual-action consistency. A deterministic router infers the disturbance type and recoverability state from runtime evidence without access to the injected disturbance label. Under the bounded settings considered here, finite delay, localized overlays, and pre-execution action drift may retain sufficient evidence for guarded recovery, whereas persistent replay becomes unrecoverable once stale-frame evidence activates the hold latch.

\subsection{Action-Conditioned Foveation and Consistency Monitoring}
\label{sec:foveation-monitoring}
ActFovea operationalizes this principle by defining the fovea as a dynamically updated preserve region anchored to projected contact and motion. Its location follows robot kinematics and action history, so the high-fidelity region moves with the predicted interaction rather than remaining fixed in image coordinates.

For each enabled camera $v$, ActFovea estimates a projected contact center $c_t^v$ and a short motion corridor $\Gamma_t^v$. With calibrated joints, current and action-extrapolated configurations are mapped by forward kinematics to gripper pinch centers and projected through the camera model. When strict projection is unavailable, the implementation uses constant-velocity Cartesian extrapolation or the previous tracked center together with an image-plane direction inferred from joint changes, velocity, and the previous action.

Let $M_{c,t}^v$ be a disk of radius $r_c$ centered at $c_t^v$, and let $M_{\Gamma,t}^v$ be the radius-$r_\Gamma$ corridor around the projected polyline $\Gamma_t^v$. The task-relevant preserve mask is
\begin{equation}
M_t^v=\operatorname{Dilate}
(M_{c,t}^v\lor M_{\Gamma,t}^v,r_m),
\label{eq:foveated-mask}
\end{equation}
where $r_m$ is a safety margin. Its complement defines the editable background mask. The implementation applies exponential moving-average smoothing and thresholding to obtain $\bar B_t^v$. Denoting background normalization, Gaussian smoothing, and color-grayscale mixing by $\mathcal E$, the filtered image is
\begin{equation}
\begin{aligned}
\widetilde I_t^v={}(1-\alpha\bar B_t^v)\odot I_t^v +\alpha\bar B_t^v\odot\mathcal E(I_t^v),
\end{aligned}
\label{eq:foveated-edit}
\end{equation}
where $\alpha$ is a bounded edit strength. This operation preserves the contact neighborhood and motion corridor while desensitizing the surrounding background.

The same reference supports consistency monitoring. An observed center $\hat c_t^v$ is estimated from local template matching, motion centroids, and appearance centroids. Geometric consistency decays with the distance between $\hat c_t^v$ and the projected contact center, while support from the projected corridor contributes to task consistency. Dynamic consistency combines directional and magnitude agreement between predicted and observed image displacement. Temporal evidence combines timestamp health, expected-but-missing local motion, lag estimated by short-history matching, and global replay similarity. Additional components measure motion relevance inside $M_t^v$ and action-proprioception agreement.

The implementation forms a fixed weighted mean of the available component scores, contracts low-confidence estimates toward $0.5$, and aggregates valid cameras using health-dependent weights. Let $r_t$ denote one minus the camera-weighted consistency and let $\bar r_t$ be its exponential moving average. The final risk is
\begin{equation}
R_t=\operatorname{clip}\bigl(
\beta\bar r_t+(1-\beta)r_t+
p_t^{\mathrm{cam}}+p_t^{\mathrm{lag}}+p_t^{\mathrm{cal}},0,1\bigr),
\label{eq:risk-score}
\end{equation}
where $\beta$ and the component weights are fixed implementation constants. Here, $p_t^{\mathrm{lag}}$ captures estimated lag, while $p_t^{\mathrm{cam}}$ and $p_t^{\mathrm{cal}}$ are auxiliary sensor-health penalties for unavailable camera evidence and calibration inconsistency. A deterministic router combines the resulting evidence with temporal persistence: direct lag supports temporal delay, dynamic and proprioceptive disagreement supports action drift, and strong stale/replay evidence with an active hold latch yields an unrecoverable freeze-like state. Weak or conflicting evidence is routed to consistent or unknown states.

The shared monitor produces a common risk state that can be strengthened by threat-conditioned evidence. For the evaluated localized overlay, ActFovea proposes a novel bounded region directly from the runtime images and accepts it as spatial evidence only when its onset, position, and extent remain stable across time and, when available, camera views. This evidence updates the same recoverability state used by the router. A temporary hold prevents uncertain observations from reaching the policy while the evidence is confirmed.

\begin{table*}[!t]
    \centering
    \small
    \renewcommand{\arraystretch}{1.0}
    \begin{tabular*}{\textwidth}{@{\extracolsep{\fill}}lcccccc@{}}
        \toprule
        Scenario & \multicolumn{2}{c}{Undisturbed SR} &
        \multicolumn{2}{c}{Disturbed SR} &
        Gain $\uparrow$ & NRR $\uparrow$ \\
        \cmidrule(lr){2-3}\cmidrule(lr){4-5}
        & Base VLA & ActFovea & Base VLA & ActFovea &
         & \\
        \midrule
        Action Drift  & 92.7 & 93.0 & 83.1 & \textbf{90.1} &
        $\mathbf{+7.0}$ & \textbf{73.1} \\
        Visual Delay  & 92.6 & 93.2 & 76.2 & \textbf{86.0} &
        $\mathbf{+9.8}$ & \textbf{59.8} \\
        Visual Overlay & 93.0 & 93.8 & 49.3 & \textbf{90.3} &
        $\mathbf{+41.0}$ & \textbf{93.7} \\
        \bottomrule
    \end{tabular*}
    \caption{Main recovery results averaged over the four LIBERO suites. Success Rate and NRR are reported in percent, and Gain is reported in percentage points. Each row uses its corresponding undisturbed controls.}
    \label{tab:main-results}
\end{table*}

\subsection{Consistency-Guided Recovery and Action Verification}
\label{sec:recovery-verification}
Threat typing determines what evidence should be corrected, but every instantiated candidate rejoins the same policy-verifier path. For recoverable or borderline evidence, the candidate bank retains raw and foveated observations and conditionally adds a temporally stabilized candidate for finite visual lag. The raw path avoids unnecessary modification under uncertain detections, and the foveated path uses Eq.~\eqref{eq:foveated-edit}. For finite visual delay, ActFovea translates the observation toward the kinematic reference using the observed-to-predicted center bias, with a gain that increases with estimated lag and decreasing freshness. Wrist-camera translations are attenuated. This deterministic image translation is instantiated only when the corresponding temporal evidence is present. Action-chunk drift generally retains trustworthy visual input and is therefore handled primarily through action verification and bounded execution rather than image reconstruction.

Confirmed localized overlays activate a spatial restoration operator within the same candidate bank. The last clean reference is aligned to the current image using the median dense optical flow outside the detected box. Given the overlaid region $P_t^v$, overlay pattern estimate $\widehat{Q}_t^v$, and observation-derived blending estimate $\widehat{\alpha}_t^v$, the clean region is reconstructed as
\begin{equation}
\widehat{X}_t^v=
\operatorname{clip}\!\left(
\frac{P_t^v-\widehat{\alpha}_t^v\widehat{Q}_t^v}
{1-\widehat{\alpha}_t^v},0,255
\right).
\label{eq:overlay-deblend}
\end{equation}
The first detected overlaid frame is paired with the most recent reference retained from the clean initialization buffer after background alignment. Their relationship provides the image-derived estimates $\widehat{Q}_t^v$ and $\widehat{\alpha}_t^v$. Gaussian feathering blends the reconstructed region at its boundary. A repair is admitted only if every detected view is reconstructed. The weakest-view quality combines detection confidence, clipping ratio, and boundary continuity, and it must pass the admission threshold.

The highest-priority candidate is queried first. Search expands when suspicious evidence and initial verification justify additional policy calls. Expansion is suppressed for clean-like, low-risk ambiguous, and unrecoverable evidence; an unrecoverable hold may directly emit a guard chunk. For each evaluated candidate, the verifier excludes the final gripper dimension and measures first-action direction, endpoint direction, motion magnitude, smoothness, horizon, and chunk drift. Let $u_k$ collect the corresponding agreement scores and complements of normalized penalties. The verification score is
\begin{equation}
V_k=\operatorname{clip}(w^\top u_k+b_k,0,1),
\label{eq:verification-score}
\end{equation}
where $w$ is a fixed nonnegative weight vector and $b_k$ is a threat-conditioned bonus. Chunk drift combines the lateral-to-total displacement ratio, endpoint-direction disagreement, and path curvature. Selection maximizes a utility combining $V_k$, execution mode, candidate priority, and small threat-dependent preferences.

When an overlay repair passes the admission gate, candidate expansion compares the raw and deblended observations. If their first three motion actions agree in direction and root-mean-square deviation, the raw action is retained. Otherwise, the repaired action is selected only when reconstruction evidence is strong and its verification score remains consistent with the raw action. The raw action remains the conservative fallback. During confirmed overlay recovery, the selected chunk also passes a configured per-policy action envelope before final verification, while the gripper command remains unchanged. All selection rules are fixed at inference time, and ActFovea adds neither a training objective nor parameter optimization.

\begin{figure*}[!t]
    \centering
    \includegraphics[width=\textwidth]{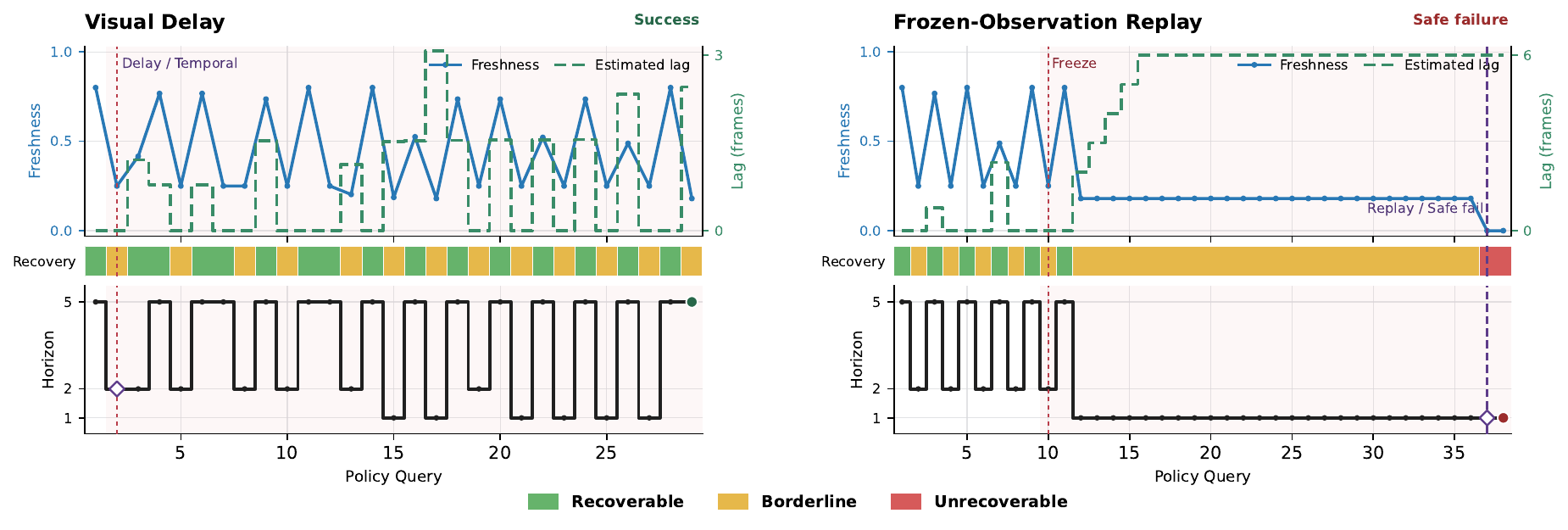}
    \caption{Representative runtime decision traces under visual delay and frozen-observation replay. Finite delay preserves recoverable temporal evidence and supports adaptive replanning, while persistent replay exhausts fresh visual evidence, shortens execution, and terminates in safe failure.}
    \label{fig:delay-freeze-timeline}
\end{figure*}

\begin{table*}[!t]
    \centering
    \small
    \renewcommand{\arraystretch}{1.0}
    \begin{tabular*}{\textwidth}{@{\extracolsep{\fill}}lcccc@{}}
        \toprule
        Method & \multicolumn{4}{c}{Success Rate (\%) $\uparrow$} \\
        \cmidrule(l){2-5}
        & Undisturbed & Action Drift & Visual Delay & Visual Overlay \\
        \midrule
        Base VLA              & 93.0 & 83.1 & 76.2 & 49.3 \\
        Action Clip/Smoothing & 82.2 & 70.4 & 70.2 & 30.9 \\
        Fixed Short Horizon   & 91.7 & 89.9 & 70.7 & 32.4 \\
        Timestamp-Only Hold   & 93.1 & 84.9 & 0.0 & 48.5 \\
        \textbf{ActFovea}     & \textbf{93.8} & \textbf{90.1} &
        \textbf{86.0} & \textbf{90.3} \\
        \bottomrule
    \end{tabular*}
    \caption{Success rates for training-free runtime methods.}
    \label{tab:runtime-baselines}
\end{table*}

\subsection{Risk-Adaptive Execution and Safe Failure}
\label{sec:risk-execution}
Execution uses two-stage arbitration. First, the consistency monitor determines whether to preserve execution, damp motion, shorten the executable horizon, or hold, producing a monitor scale $\lambda_t^{\mathrm{mon}}$ and horizon $h_t^{\mathrm{mon}}$. Second, action verification chooses normal execution, short-horizon execution, servo recovery, or safe failure, with optional caps $\lambda_t^{\mathrm{ver}}$ and $h_t^{\mathrm{ver}}$. The combined decision is
\begin{equation}
\begin{aligned}
\hat a_{t,i}^{\mathrm{mot}}
&=\lambda_t^{\mathrm{mon}}\lambda_t^{\mathrm{ver}}
a_{t,i}^{\star,\mathrm{mot}}, \qquad i<h_t,\\
h_t&=\min(h_t^{\mathrm{mon}},h_t^{\mathrm{ver}}),
\end{aligned}
\label{eq:bounded-execution}
\end{equation}
where an absent verifier cap denotes unit scale and the full chunk horizon. Monitor-level holding and verifier-level safe failure override the remaining modes and set motion to zero. When neither stage intervenes, the full chunk is preserved, while motion damping, short-horizon execution, and servo recovery reduce motion magnitude, horizon, or both. Recoverable execution preserves the gripper dimension, while holding carries forward the previous gripper command. The runtime client re-queries the policy after $h_t$ actions.

Finite visual delay preserves an ordered history that can support temporal alignment, whereas persistent frozen-observation replay does not. When stale content persists despite image-plane alignment, ActFovea switches from recovery to conservative execution. It latches hold after immediate strong replay evidence or a configurable stale streak and releases it after sufficient fresh evidence. Hold-triggering or high-risk fail-closed responses may prepend one clipped reverse action before filling the remaining chunk with holds. Here, safe failure denotes conservative motion suppression. Formal collision-avoidance guarantees lie outside the scope of this mechanism.

\section{Experiments}

\subsection{Experimental Setup}

\paragraph{Benchmark and Protocol.}
We evaluate ActFovea with the frozen LIBERO \cite{liu2023libero} checkpoint of $\pi_0$ on four ten-task suites: LIBERO-Spatial, LIBERO-Object, LIBERO-Goal, and LIBERO-10. Each task uses 50 episodes, yielding 2,000 episodes per method-scenario cell. All comparisons use the same checkpoint and matched task and execution configurations; ActFovea is applied only at inference time.

\paragraph{Disturbances.}
Smooth Action-Chunk Drift uses task-family-specific phase templates to perturb the motion part of an action chunk over a phase-conditioned window. Multi-View Visual Feedback Delay holds both exterior and wrist views three frames behind the current proprioceptive state. Persistent Localized Visual Overlay blends fixed localized checker patterns into both camera streams after a short clean warm-up. Frozen-Observation Replay reuses its trigger frame in both views until termination. ActFovea receives only the resulting observations and standard runtime inputs; disturbance labels and generator parameters are not provided.

\paragraph{Metrics.}
We report task success rate (SR), absolute defense gain in percentage points, and normalized recovery rate (NRR):
\begin{equation}
\begin{aligned}
\mathrm{Gain}&=S_{\mathrm{D+AF}}-S_{\mathrm{D}},\\
\mathrm{NRR}&=
\frac{S_{\mathrm{D+AF}}-S_{\mathrm{D}}}
{S_{\mathrm{C}}-S_{\mathrm{D}}}\times 100\%.
\end{aligned}
\label{eq:recovery-metrics}
\end{equation}
Here $S_{\mathrm{C}}$, $S_{\mathrm{D}}$, and $S_{\mathrm{D+AF}}$ denote Base clean, Base disturbed, and ActFovea disturbed success. For frozen-observation replay, each episode is assigned to Task Success, Timely Safe Failure, or Unprotected Failure. Timely Safe Failure denotes reaching the terminal safe-failure state within the detection and action budgets, with all safe-failure actions inside the executor bound; Unprotected Failure covers the remaining unsuccessful episodes.

\subsection{Recovery across Spatial, Temporal, and Action Disruptions}
\label{sec:recovery-results}

We first ask whether the same safeguarding loop transfers across spatial visual corruption, finite temporal misalignment, and action-trajectory drift. Table~\ref{tab:main-results} shows consistent recovery across all three disturbance axes. Persistent visual overlays produce the largest controlled loss, reducing Base VLA success from 93.0\% to 49.3\%. ActFovea restores success to 90.3\%, a gain of 41.0 percentage points that recovers 93.7\% of the induced loss.

Under finite visual delay, ActFovea raises success from 76.2\% to 86.0\%, a gain of 9.8 points that recovers 59.8\% of the performance loss caused by the delay. Under action drift, it raises success from 83.1\% to 90.1\%, a gain of 7.0 points that recovers 73.1\% of the disturbance-induced performance loss. Across all three evaluations, ActFovea preserves performance under undisturbed operation. Figure~\ref{fig:delay-freeze-timeline} further contrasts the runtime decisions induced by finite delay and persistent replay. Delay retains intermittent freshness and bounded lag estimates, allowing recovery and adaptive replanning. Persistent replay instead removes fresh evidence, drives the response from recovery to an unrecoverable state, and reduces the executable horizon before safe failure.

\begin{table}[t]
    \centering
    \footnotesize
    \renewcommand{\arraystretch}{1.0}
    \begin{tabular*}{\columnwidth}{@{\extracolsep{\fill}}lccc@{}}
        \toprule
        Method & \shortstack{Task\\Success} &
        \shortstack{Timely Safe\\Failure} &
        \shortstack{Unprotected\\Failure} \\
        \midrule
        Base VLA            & 3.05 & 0.00 & 96.95 \\
        Timestamp-Only Hold & 0.00 & 0.00 & 100.00 \\
        w/o Hold/Safe-Fail  & 0.65 & 0.00 & 99.35 \\
        \textbf{ActFovea}   & \textbf{0.00} & \textbf{100.00} &
        \textbf{0.00} \\
        \bottomrule
    \end{tabular*}
    \caption{Outcome rates under frozen-observation replay, showing whether each method completes the task, reaches timely safe failure, or remains unprotected.}
    \label{tab:freeze-safety}
\end{table}

\begin{table*}[!t]
    \centering
    \small
    \renewcommand{\arraystretch}{1.0}
    \begin{tabular*}{\textwidth}{@{\extracolsep{\fill}}lcccccc@{}}
        \toprule
        Variant & \multicolumn{2}{c}{Action Drift} &
        \multicolumn{2}{c}{Visual Delay} &
        \multicolumn{2}{c}{Visual Overlay} \\
        \cmidrule(lr){2-3}\cmidrule(lr){4-5}\cmidrule(lr){6-7}
        & Success Rate & Gain & Success Rate & Gain &
        Success Rate & Gain \\
        \midrule
        w/o Threat Typing
            & 87.5 & $+4.4$ & 80.5 & $+4.3$ & 41.7 & $-7.6$ \\
        w/o Recovery Bank
            & 87.5 & $+4.4$ & 84.0 & $+7.8$ & 16.0 & $-33.3$ \\
        w/o Candidate Expansion
            & 90.6 & $+7.5$ & 85.4 & $+9.2$ & 17.6 & $-31.7$ \\
        w/o Action Verification
            & 81.9 & $-1.2$ & 78.5 & $+2.3$ & 92.1 & $+42.8$ \\
        \textbf{Full ActFovea}
            & \textbf{90.1} & $\mathbf{+7.0}$ &
            \textbf{86.0} & $\mathbf{+9.8}$ &
            \textbf{90.3} & $\mathbf{+41.0}$ \\
        \bottomrule
    \end{tabular*}
    \caption{Component ablations under the three recoverable disturbances. Success Rate is measured under the indicated disturbance. Every Gain is computed relative to the corresponding disturbed Base VLA result in Table~\ref{tab:main-results}.}
    \label{tab:ablation}
\end{table*}

\subsection{Training-Free Runtime Comparisons}
\label{sec:runtime-comparison}

The comparison methods isolate common explanations for runtime robustness. Action Clip/Smoothing applies a fixed bounded-smoothing rule to the motion dimensions while leaving the gripper command unchanged. Fixed Short Horizon replans after a shortened action prefix. Timestamp-Only Hold suspends execution using frame-age and repeated-image evidence and releases after fresh observations return.

Table~\ref{tab:runtime-baselines} shows that Fixed Short Horizon reaches 89.9\% under action drift, closely approaching ActFovea's 90.1\% and confirming that frequent replanning is well suited to this disturbance. Its fixed response does not transfer across inconsistency types: visual-delay and visual-overlay success are 70.7\% and 32.4\%. Action Clip/Smoothing reduces undisturbed success to 82.2\% and remains below the corresponding Base VLA result under all three disturbances. Timestamp-Only Hold preserves undisturbed behavior and reaches 48.5\% under the visual overlay, but continuous visual delay keeps its timestamp rule active and reduces success to zero. ActFovea is the only evaluated runtime method that preserves undisturbed performance while improving all three recoverable disturbances. Its advantage therefore comes from threat-conditioned recovery and verification rather than a single conservative action rule.

\subsection{Bounded Safe Failure under Frozen-Observation Replay}
\label{sec:freeze-results}
We next test the temporal recoverability boundary. Finite visual delay retains an ordered history, whereas persistent replay removes the fresh evidence required for task recovery. The target in the latter case is therefore a bounded terminal outcome rather than task completion.

As Table~\ref{tab:freeze-safety} shows, ActFovea prioritizes a bounded terminal outcome over task completion when fresh evidence is unavailable. All 2,000 frozen-replay episodes reach timely safe failure, with no task successes or unprotected failures. Safe fail occurs immediately after replay detection; only 2.0 bounded action steps are executed afterward, with a mean cumulative action-space motion norm of 0.326 and no action-bound violations. Without Hold/Safe-Fail, the controller continues for 259.2 actions after detection and accumulates a motion norm of 241.98. The full executor therefore reduces post-detection action count by 99.23\% and cumulative motion by 99.87\%.

Timestamp-Only Hold spends 96.55\% of frozen-replay queries holding but never converts that state into terminal safe failure. This distinction explains why sustained zero motion alone is not equivalent to a bounded failure protocol.

\subsection{Mechanism Analysis}
\label{sec:ablation-runtime}

\begin{figure}[!t]
    \centering
    \includegraphics[width=\columnwidth]{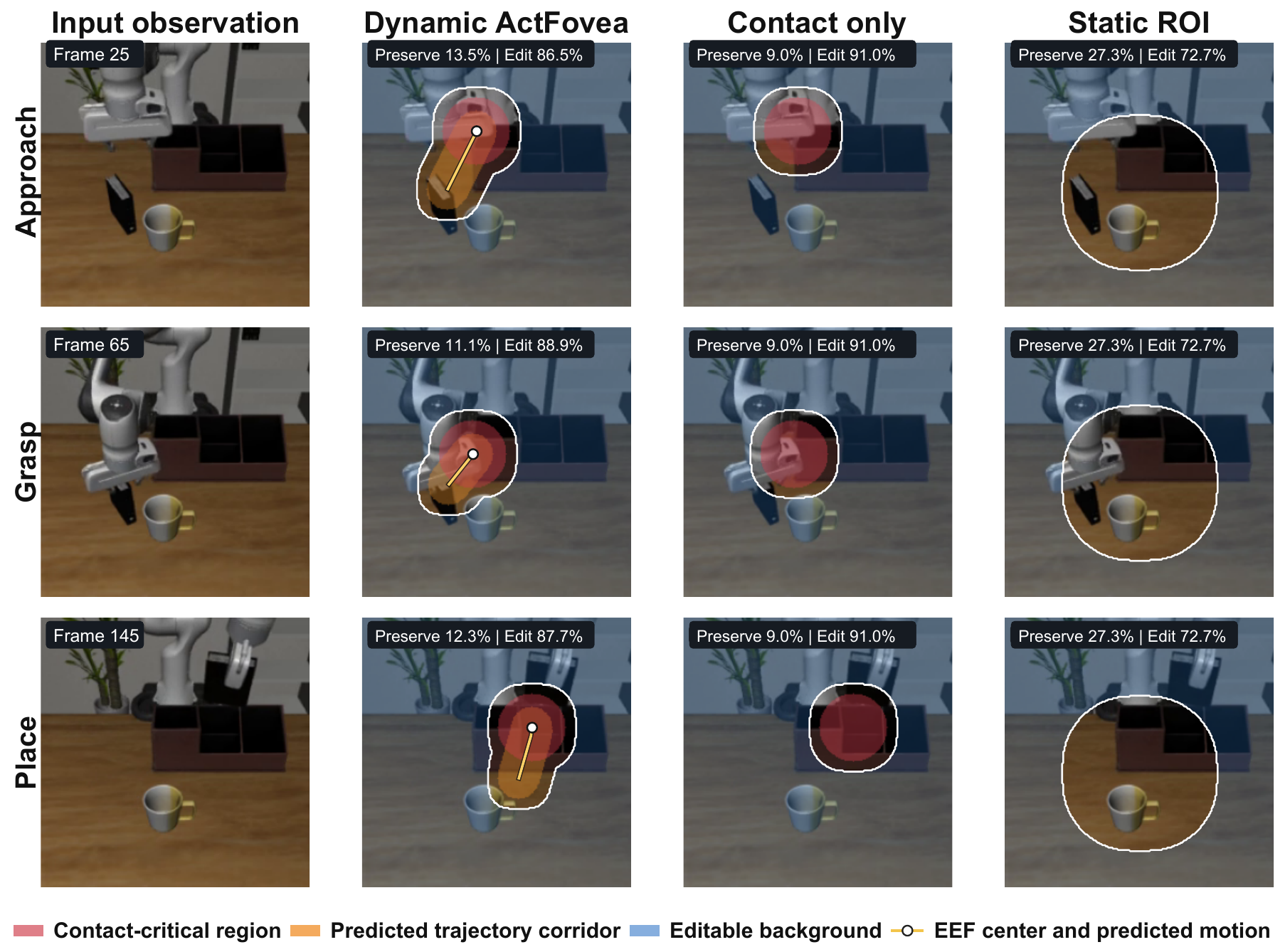}
    \caption{Qualitative foveation comparison across task phases. Dynamic ActFovea tracks the contact-critical region and predicted motion, whereas contact-only and static regions omit motion support or phase-dependent adaptation.}
    \label{fig:dynamic-fovea-comparison}
\end{figure}

Figure~\ref{fig:dynamic-fovea-comparison} illustrates the spatial mechanism across approach, grasp, and place phases. The dynamic fovea follows the contact-critical region while extending along predicted motion. A contact-only region omits this motion corridor, whereas a static region cannot adapt its spatial support as the interaction evolves.

Table~\ref{tab:ablation} presents a common success-rate and gain view across the three recoverable disturbances, using the disturbed Base VLA results in Table~\ref{tab:main-results} as a shared reference. Full ActFovea provides gains of 7.0 points under Action Drift and 9.8 points under Visual Delay. Bypassing action verification reduces these gains to $-1.2$ and $+2.3$ points, respectively, showing that trajectory validation is the main mechanism that turns temporal and action-side candidates into reliable recovery. Threat typing is especially important for Visual Delay, where routing all evidence to the unknown state reduces the gain to 4.3 points.

Under Visual Overlay, removing threat typing, the recovery bank, or candidate expansion changes the full-system gain of 41.0 points to $-7.6$, $-33.3$, and $-31.7$ points, whereas the variant without action verification retains a 42.8-point gain. Spatial recovery therefore relies primarily on locating the corrupted region and constructing a usable observation alternative, while action verification provides the shared conservative acceptance gate. The resulting division of labor allows one pipeline to support spatial, temporal, and action-side disruptions without applying every intervention uniformly.

\FloatBarrier

\section{Conclusion}
We presented ActFovea, a policy-interface-based runtime safeguard that treats spatial corruption, temporal misalignment, and action-trajectory drift as violations of spatiotemporal visual-action consistency. Action-conditioned foveation preserves interaction-relevant evidence, the consistency monitor determines whether recovery remains justified, and disturbance-conditioned candidate observations are verified before bounded execution, while the underlying VLA policy remains frozen throughout. When fresh perceptual grounding is lost, the same safeguarding loop transitions from recovery to safe failure. Across multiple LIBERO tasks with $\pi_0$, ActFovea restores visual-overlay success from 49.3\% to 90.3\%, improves visual-delay and action-drift success by 9.8 and 7.0 percentage points, and preserves performance under undisturbed operation. Under frozen-observation replay, all evaluated episodes reach timely safe failure with no unprotected failures. Ablations reveal complementary roles: observation recovery enables spatial restoration, while action verification is central to temporal and action-side recovery. These results support spatiotemporal visual-action consistency as a practical basis for training-free runtime VLA safeguarding.

\bibliography{aaai2027}

\end{document}